# Harnessing AI Agents to Advance Research on Refugee Child Mental Health

Aditya Shrivastava[1][0009-0003-5733-3260] Komal Gupta[2][0009-0008-3929-1710]

and Shraddha Arora[3][0000-0003-2209-9278]

[1] The Governor's Academy, Byfield, MA, USA
[2,3] The Northcap University, Haryana, India
Komalgupta991000@gmail.com

**Abstract.** The international refugee crisis deepens, exposing millions of displaced children to extreme psychological trauma. This research suggests a compact, AI-based framework for processing unstructured refugee health data and distilling knowledge on child mental health. We compare two Retrieval-Augmented Generation (RAG) pipelines, Zephyr-7B-beta and DeepSeek R1-7B, to determine how well they process challenging humanitarian datasets while avoiding hallucination hazards. By combining cutting-edge AI methods with migration research and child psychology, this study presents a scalable strategy to assist policymakers, mental health practitioners, and humanitarian agencies to better assist displaced children and recognize their mental wellbeing. In total, both the models worked properly but significantly Deepseek R1 is superior to Zephyr with an accuracy of answer relevance 0.91

**Keywords:** Retrieval-Augmented Generation, Zephyr-7B-beta, DeepSeek R1-7B, Answer Relevance, Hallucination, LLM as a Judge, Refugee Crises

## 1. Introduction

The worldwide refugee crisis has lately heightened, posing serious humanitarian challenges. Among these, the psychological impact on displaced children still constitutes one of the least explored yet most imperative issues. Refugee children tend to suffer profound trauma as a result of forced displacement, loss of domicile, and insecurity in refugee camps. Prompt detection and intervention of their mental health issues are equally imperative for successful humanitarian response and policy development. The growing frequency and magnitude of forced displacement events, caused by conflict, political unrest, and climate change, have resulted in a record global refugee crisis. By the end of 2023, more than 43.3 million children had been forcibly displaced, as reported by the United Nations High Commissioner for Refugees (UNHCR) [1]. Such children usually experience extreme psychological distress because of violence, separation from family members, and the uncertainty of refugee life. Research shows that displaced children have higher chances of Post-Traumatic Stress Disorder (PTSD), depression, and anxiety disorders, but there is limited comprehensive and systematic examination of their mental health.

One of the main challenges in research in refugee mental health is data access and interpretation. Health records and psychological questionnaires of refugees tend to be unstructured, disjointed, and not easily standardizable for large studies. Moreover, sensitive data analyses need models that reduce hallucinations—a prevalent problem in LLMs where AI provides deceptive or wrong outputs. As a result of such challenges, the paper suggests an AI-driven framework for extracting and aggregating information across various sources such as policy briefs, health evaluations, and social media narratives to offer dependable insights into refugee children's mental health conditions.

To accomplish this, the suggested study contrasts two AI models based on Retrieval-Augmented Generation (RAG) methods. This setup enables us to experiment with various AI architectures' management of sensitive humanitarian data, retrieval process optimization, and hallucination risk management. The suggested research offers a foundational method for incorporating AI into international migration studies, supporting policy development and intervention measures for displaced children.

## 2. Literature Review

Artificial intelligence (AI) is being applied in many fields of healthcare, such as oncology, radiology, and dermatology. Its application in mental health care [3] and neurobiological studies has been modest otherwise (Lee et al., 2021) [4]. With the mounting burden of psychiatric illnesses and a severe lack of mental health care professionals, AI can potentially contribute to a revolution in the detection of high-risk individuals and creating timely interventions. Although published evidence of AI application in neuropsychiatry is scarce, recent studies have proven the utility of AI in processing electronic health records, brain imaging, sensor-based monitoring devices, and social media sites to predict and categorize mental disorders, such as suicidality.

Recent developments in AI technologies for mental health have proven its applicability in fields like teletherapy, diagnostic modeling, and patient behavior monitoring in real-time. Olawade et al. (2024) identify the potential of AI to improve mental health services through early detection, customized treatment plans, and virtual therapists powered by AI [5]. AI has also been very useful in addressing accessibility issues and bringing mental health services closer to underserved communities, such as refugees.

AI-driven telehealth interventions have become vital instruments in treating mental health issues among vulnerable groups, including displaced children due to the refugee crisis. Mansoor and Ansari (2025) compared SAMHSA data and determined that AI-driven telehealth therapy was as effective as face-to-face therapy for treating depression and more effective for anxiety disorders. Their research found key predictors of telehealth success, such as session frequency, previous diagnosis, and socioeconomic status, highlighting the need for personalized AI-based mental health therapy [6].

The use of large language models (LLMs) in refugee mental health care is marred by a number of challenges that need to be overcome to ensure their optimal effectiveness. One of the major challenges is their poor understanding of the multicultural backgrounds of the refugees, which can result in confusion and ineffectiveness of response. To counter this, fine-tuning LLMs with culture sensitivity training and context-specific knowledge bases through methods such as transfer learning and supervised fine-tuning is necessary. Additionally, access barriers such as prohibitive internet costs, absence of devices, and low digital literacy among refugees also impede their use. Creation of a low-bandwidth, voice-based LLM interface especially for low literacy groups, along with programs for the distribution of low-cost or refurbished devices, may make them more useful. Additionally, since mental health care entails enormous knowledge of the state of an individual, LLMs alone may be insufficient [7].

The Integration of large language models (LLMs) in mental health studies on social media platforms has provided new avenues for the identification and detection of mental health diseases using user-generated content. The conventional discriminative models do not possess the capacity to generalize and interpret effectively, thus compromising the performance overall. In addressing these problems, Jahani, S., Dehghanian, Z., & Takian, A. (2024) performed a study of MentaLLaMA, an open-source instruction-following LLM to extend the interpretability of mental health evaluations. MentaLLaMA outperforms current models by offering human-readable, elaborate explanations in addition to predictions. The development of MentaLLaMA points to the necessity of designing LLMs for mental health applications, especially improving interpretability and trustworthiness for both clinical and digital mental health interventions [8].

Zhijun Gua and Alvina Lai critically assess the use of LLMs in mental health, with a focus on their suitability and efficacy in early detection, digital treatment, and clinical uses. LLMs have been used to assess social media posts for detection of mental health disorders, support AI-based therapy chatbots, and help clinicians through documentation and treatment planning automation. As much as evidence suggests their usefulness in the identification of vulnerable groups and the provision of scalable mental health interventions, bias, data privacy, interpretability, and ethical concerns limit their widespread use. The "black-box" nature of LLMs raises clinical integration challenges, as clinicians require transparent and reliable AI-driven assessments. Through enhanced methodologies and interdisciplinary collaboration between AI researchers and clinicians, LLMs can play an important role in bringing about a transformative role in global mental health outcomes [9].

Together, these works show the growing capability of agentic AI to reshape mental health scholarship by making it possible for more nuanced and advanced understanding of mental health dynamics. This paper takes inspiration from these seminal works by using a highly sophisticated LLM model to study refugee migration flows and their subsequent mental health impacts, aiming to fill an important gap in the literature and make actionable suggestions to policymakers and practitioners.

This paper is among the first to comparatively benchmark Retrieval-Augmented Generation (RAG) pipelines, specifically Zephyr-7B-beta and DeepSeek R1-7B, for retrieving insights from unstructured refugee mental health reports. Unlike previous works discussing the overall potential of AI use in healthcare or mental health, this work pins down its scope to the specific overlap between AI designs and humanitarian datasets regarding displaced children. Our innovation is in three directions: designing a culture and context-aware assessment system that quantifies hallucination risk and answer suitability in terms of LLM-based assessment tools like GPT-4, reconfiguring data pipelines to navigate culturally and contextually rich refugee documents based on semantic-aware chunking and maximal marginal relevance (MMR), and designing a benchmark model exclusively for refugee children's mental health data. This integrated, replicable design is the first to offer both quantitative indicators and qualitative ratings, establishing a methodological benchmark for subsequent applications of AI in research among vulnerable populations.

## 3. Problem Formulation & Objectives

**The proposed method aims to address three main issues in the application of artificial intelligence in humanitarian research:**

1. Unstructured refugee health documents extraction and structuring – Most refugee health reports are in non-standardized formats, and therefore they are hard to analyze using traditional AI models.
2. The Research AI Agent is specifically trained to answer only research questions related to refugee crises, thus ensuring that its answers are relevant, targeted, and ethically sound.
3. Preventing hallucination threats in domain-specific LLM use cases – Since humanitarians' data is vulnerable to manipulation, preserving the fact-based integrity of AI-recommended insights is paramount.

3.1 Primary Research Objectives

This research is motivated by two significant issues in humanitarian artificial intelligence research. First, to collect efficient information extraction from unstructured refugee health records. The records contain complicated and context-dependent information that must be parsed and analyzed correctly to provide mental health diagnoses. Second, the research aims to reduce hallucination risks in applications of domain-specific large language models. Hallucinations are serious risks in delicate contexts such as refugee mental health research, where errors can cause inappropriate interventions.

The main goals of this work are to create a comparative framework for the assessment of LLM performance for refugee mental health research and to modify RAG pipelines for processing sensitive psychosocial data. Furthermore, the research seeks to create end-to-end benchmarks for hallucination control that are modified for humanitarian settings. Through the accomplishment of these goals, this framework enhances the validity and the interpretability of AI-driven insights in a very sensitive application domain.

## 4. Methodology

In this research study, a systematic strategy is adopted to aimed at solving two key issues: efficient information extraction from unstructured refugee health records and reducing the dangers of hallucinations in the use of LLM in domains. The suggested strategy is organized into four main components: data acquisition, model development and RAG implementation, domain-specific customization, and an evaluation framework leveraging LLM-based assessment. Figure 1 illustrates a more structural flow of the desired approach.

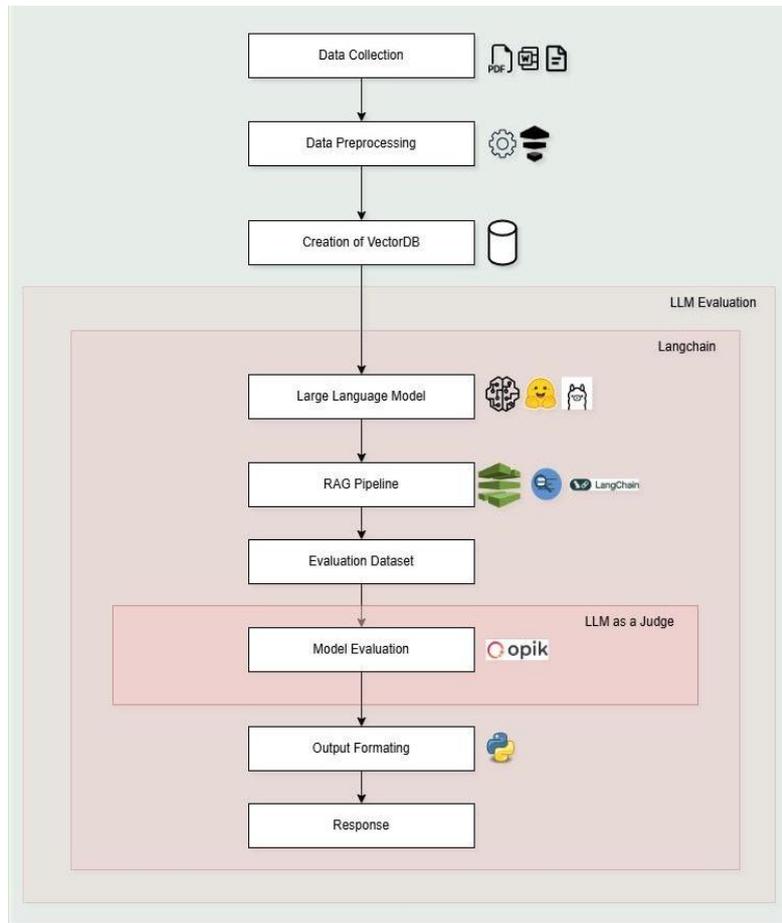

**Fig. 1.** Architecture diagram of the proposed approach.

### 4.1 Data Acquisition

The first step of data collection for this study is the meticulous gathering of high-quality data from a wide range of sources. A corpus is established that consists of government migration reports, UNHCR and other NGOs publications, anonymized medical records of medical centers, and relevant social media posts [2]. This data set, obtained from various sources, is gathered with utmost caution in strict adherence to ethical standards, including GDPR, to ensure the privacy and security of sensitive information. The diversity of the data allows the refugee crisis to be represented in an all-inclusive manner with particular focus on the determinants of child mental health, thus allowing a strong platform for the development of future models.

### 4.2 Model Development and RAG Implementation

In establishing contextually adequate and factually accurate outputs, the suggested model framework is put on a Retrieval-Augmented Generation (RAG) model architecture. In this architecture, external knowledge retrieval is incorporated into the text generation process, thus grounding the model outputs on up-to-date and reliable information. The framework seeks to weigh the computational expense of the system and task complexity cautiously with the utilization of a model configuration suitable to the available resources. By doing this, the conversion of unstructured text to meaningful representations and recovery of relevant contextual information are facilitated, which, in turn, aid the generative process. Through the structuring of the system in this way, minimized the possibility of producing errors while maintaining the depth and accuracy required for sensitive research in the humanitarian field.

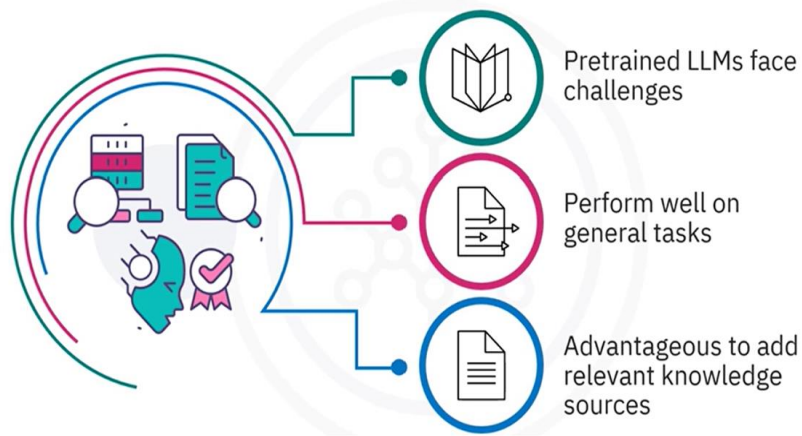

**Fig. 2.** Overview of the RAG process integrating external knowledge for enhanced model performance.

4.3 Domain-Specific Customization

Recognizing the need for a balanced understanding of precise contextual hints in refugee mental health studies, the proposed model is enriched by making domain-specific modifications. This involves adapting the base model through specially tailored training with a dataset that includes migration stories, psychological assessments, and relevant background history. Specific training of this sort enhances the ability of the model to detect and understand linguistic subtleties and cultural frames of reference within refugee discourse. Additionally, the tailoring enhances the accuracy of the output while ensuring that the model is well-adapted to the special character of psychosocial knowledge in humanitarian contexts.

4.4 Evaluation Framework and LLM Assessment

A critical component of the methodology presented is a comprehensive evaluation mechanism designed to rigorously quantify model performance. Multi-perspective evaluation methodology is employed that accounts for accuracy and relevance of the produced content with respect to the questions being posed. Towards this, OpenAI GPT-4 was used as an evaluation authority, which employs a standardized rubric to measure primary measures—i.e., hallucination rate and answer relevance. This method allows us to obtain objective, quantitative measures of model output quality along with the use of qualitative feedback. Combined approach of automated test tools and expert human assessment is performed, which allows the system to continuously learn and improve, thereby ensuring that it is always committed to high standards of reliability and ethical integrity in the evaluation of refugee children's mental health.

## 5. Experimental Design

Experimental design is centered around comparative architecture that processes input data through various steps—ranging from preprocessing to LLM evaluation—such that both retrieval accuracy and generation fidelity are comprehensively tested. Total design starts from a well-defined input dataset, which is preprocessed to be ready for embedding. The subsequent embedding step transforms the unstructured text to numerical representations, which are stored in a vector database. The representations enable an advanced retrieval mechanism, which in turn feeds the LLM processing step. Outputs are then tested against pre-defined metrics that measure both response accuracy and hallucination rates. To ensure experimental runs are consistent, the design comprises several control variables. The same document corpus comprising 16 PDF files is used throughout the study. In addition, a fixed set of 50 domain-specific queries is used, and all experiments are run at a fixed temperature setting of 0.2.

### 5.1 Hyperparameter Tuning

The selection of a temperature value of 0.2 is a deliberate choice grounded in the research focus. The temperature in LLMs controls the degree of variability and creativity in responses—higher values introduce more randomness, while lower values encourage more deterministic and consistent outputs [10]. Given that the scope of this study is to analyze the impact of refugee crises on children's mental health, the requirement for validated, accurate data with minimal hallucinations is paramount. Research conducted preliminary experiments with varying temperature values and observed that higher temperatures (e.g., 0.7 or 1.0) led to increased variability and occasional inaccuracies, which could compromise the integrity of the findings. Conversely, a lower temperature (0.2) provided the optimal balance by maintaining response consistency while ensuring sufficient adaptability in language generation.

### 5.2 LLM as a Judge

The evaluation phase uses OpenAI GPT-4 as the judge to supply unbiased ratings for the outputs. The independent variables in the suggested design include numerous embedding strategies, different text segmentation methods, and different model structures, all of which are expected to have performance effects in quantifiable manners. By setting the temperature at 0.2, we limit the possibility of over-model hallucination while ensuring coherence and factual accuracy, with the result that conclusions drawn from the study remain valid and scientific in nature.

### 5.3 Hardware Configuration

The hardware requirements used in the proposed approach are listed in Table 1, and the software used is VS Code to access the computing environment and using GPU Processor - AMD Ryzen 7 5800H with Radeon Graphics, 3201 Mhz, 8 Core(s), 16 Logical Processor(s).

**Table 1.** Hardware specifications

| | |
|---|---|
| **Operating system** | Windows 11 Home Single Language 64-bit Version: 22621.1848 |
| **Microprocessor** | Intel(R) Core (TM) i5-8265U CPU @ 1.60GHz |
| **System memory** | 8 GB |

### 5.4 Zephyr Model Workflow

#### 5.4.1 Data Preprocessing

The Zephyr pipeline starts with PDF ingestion via the PyMuPDFLoader, which effectively extracts text from structured and semi-structured documents. After extraction, the text is sent through a RecursiveTextSplitter that is set to output 500-token chunks with a 50-token overlap. This fixed-size chunking technique guarantees uniform segmentation of documents, although it does not necessarily honor the semantic or hierarchical structure in the text. Despite its simplicity, this approach is computationally efficient and well-suited to systems that must process large volumes of data quickly.

#### 5.4.2 Vector Database Setup

Following segmentation, each chunk is embedded using the all-MiniLM-L6-v2 model, which transforms text into dense vectors reflecting semantic similarity. These vectors are then stored in FAISS (Facebook AI Similarity Search), a robust library for vector-based retrieval. The choice of all-MiniLM-L6-v2 balances performance and resource demands, making it a common go-to embedding model for general-purpose tasks. Because Zephyr's chunking strategy yields uniformly sized segments, indexing and retrieval in FAISS proceed straightforwardly.

#### 5.4.3 LLM Configuration

Zephyr employs a HuggingFace endpoint for large language model inference, typically constrained by a 4,000-token context window. The decoding strategy used here is greedy search, which deterministically selects the highest-probability token at each step. While this approach yields predictable outputs, it can sometimes limit the diversity and naturalness of the generated responses. However, the uniform chunk sizes and simpler retrieval pipeline reduce overhead, making Zephyr a practical choice for scenarios that prioritize speed and consistency.

#### 5.4.4 RAG Implementation

Zephyr's Retrieval-Augmented Generation (RAG) pipeline adopts a top-k retrieval scheme (k=3), selecting the three most semantically similar segments to a user query. These segments, along with the user query, are then fed into the language model, ensuring that generated responses are anchored in relevant text chunks. Though the method does not account for the structural or hierarchical context of documents, it provides a robust, streamlined approach for straightforward tasks that demand moderate levels of contextual accuracy/

## 5.5. Deepseek R1 Workflow

### 5.5.1 Data Preprocessing

In contrast to the fixed-size splitting of Zephyr, Deepseek employs a DocumentConverter that extracts and formats text while retaining structural markers. Text is then segmented using a MarkdownHeaderSplitter, which respects headings, subheadings, and other semantic cues. This strategy produces chunks that better align with logical divisions in the source material, crucial for contexts like refugee health reports that contain layered or hierarchical information.

### 5.5.2 Vector Database Setup

In Deepseek workflow nomic-embed-text embedding modelis used to convert every semantically segmented chunk into dense embeddings. Like Zephyr, the system employs FAISS for indexing. But with the union of semantically consistent segments and a structured text-optimized model, Deepseek's vector store is more reflective of the document's intrinsic meaning and context. This structure-awareness enhances retrieval accuracy, especially when handling domain-specific, sensitive information where subtle context matters.

### 5.5.3 LLM Configuration

The Deepseek pipeline integrates with an Ollama API for its large language model backend, supporting an 8,000-token context window—double that of Zephyr. The larger window means more rich and complete context can be incorporated at inference time. On top of that, Deepseek uses a nucleus sampling (p=0.9) decoding strategy, which is capable of generating more diverse and natural-looking outputs than the determinism of greedy search. This adaptability is particularly valuable in humanitarian research, where obtaining nuanced or context-dependent information is paramount.

### 5.5.4 RAG Implementation

One of the standout differentiators is Deepseek's application of Maximal Marginal Relevance (MMR) for retrieval with k=2 and λ=0.5. Instead of merely selecting the top-k most similar chunks, MMR blends similarity with diversity, reducing redundancy in the returned segments. By doing so, the resulting context presented to the LLM will be both relevant and comprehensive, minimizing the risk of omitting important information. This is especially true for refugee mental health statistics, which can cover several subtopics or sections within one report.

## 6. Comparative Analysis: Why Deepseek Is Better?

While both Zephyr and Deepseek utilize RAG pipelines to ensure factual grounding in LLM responses, the Deepseek approach typically outperforms Zephyr in complex humanitarian research for the following reasons:

1. **Semantic-Aware Segmentation** - Deepseek's MarkdownHeaderSplitter does honor document structure, producing pieces that better capture contextual meaning. This is particularly important when processing multi-levelled refugee mental health reports, where nuance makes all the difference in interpreting psychological evaluation.
2. **Optimized Retrieval Strategy** - Maximal Marginal Relevance does not just select the most relevant chunks but also encourages diversity so that the model is not overwhelmed by repetitive or redundant information. This is a mechanism worth its weight in gold when data may be repetitive or complex, as is typical of health and policy reports.
3. **Extended Context Window and Nucleus Sampling** - With an 8k-token limit and a non-deterministic decoding strategy, Deepseek can incorporate a richer variety of information in its outputs. This is particularly beneficial for capturing nuanced psychosocial factors, historical context, or policy constraints that can be overlooked by simpler models or shorter context windows. With an 8k-token cap and non-deterministic decoding approach, Deepseek is able to include a more diverse array of information within its outputs. This is especially useful for obtaining subtle psychosocial considerations, historical background, or policy requirements that may not be captured through less complex models or narrower context windows.
4. **Domain-Specific Robustness** - By merging structured text segmentation with an embedding model that is fine-tuned for hierarchical data, Deepseek shows higher accuracy and robustness in sensitive areas. The increased context window size also reduces hallucination hazards since the LLM can draw upon a wider knowledge base during inference.

In conclusion, while Zephyr provides an efficient, low-resource solution to less complex operations, Deepseek's architecture and retrieval methods in general are well adapted to complexities and sensitivities of refugee mental health research. Through the maintaining of document structure, context diversification, and facilitation of more holistic inference, Deepseek is better qualified as the most effective option in domain-specific analysis that requires both depth and precision. A structured Tabular model comparative configuration for the model appears in Table 2.

Table 2. Hyperparameter configuration of both workflows.

| Parameter | Zephyr Model Workflow | Deepseek R1 Workflow |
|---|---|---|
| **Chunking Strategy** | 500 tokens, 50 overlap (RecursiveTextSplitter) | Header-aware splitting (MarkdownHeaderSplitter) |
| **Embedding Model** | all-MiniLM-L6-v2 | nomic-embed-text |
| **Retrieval Method** | Top-k = 3 | MMR (k=2, $\lambda$=0.5) |
| **LLM Backend** | HuggingFace Endpoint | Ollama Localhost API |
| **Context Window** | 4k tokens | 8k tokens |
| **Decoding Strategy** | Greedy search | Nucleus sampling (p=0.9) |
| **Document Preprocessing** | PyMuPDFLoader, fixed-size chunking | DocumentConverter, semantic chunking |
| **Vector Database** | FAISS | FAISS |

## 8. Results

For Evaluation, Opik Evaluation framework is combined with OpenAI GPT-4 as the deciding judge to determine the performance of the two RAG pipelines. This Evaluation was concentrated on two vital metrics: the rate of hallucinations and answer relevance of the produced answers. For the Zephyr pipeline, the hallucination metric value was 0.32, while its answer relevance value was 0.88. In contrast, the Deepseek pipeline scored 0.12 in hallucination metric and 0.92 in answer relevance as illustrated in Table 3. The scores suggest that the Deepseek method significantly minimizes the frequency of hallucinated content, as indicated by its lower hallucination metric, yet offering slightly higher answer relevance. The combination of a header-aware segmentation approach and an MMR-based retrieval process within the Deepseek pipeline is likely responsible for this enhancement, as it more effectively maintains document structure and provides varied but relevant context to the LLM. In general, the findings show that the Deepseek R1 7B pipeline performs better than the Zephyr 7B-beta setup, providing more stable and contextually correct outputs in domain-specific uses of refugee mental health studies.

Table 3. Result Analysis

| Evaluation metrics | Zephyr Model | Deepseek R1 Model |
|---|---|---|
| Answer Relevance | 0.88 | 0.91 |
| Hallucination | 0.32 | 0.12 |

## 9. Conclusion

This research methodology establishes a holistic comparative framework for assessing RAG pipelines within the context of humanitarian AI research on refugee mental health. By overcoming the difficulties of successful information extraction from unstructured documents and reducing hallucination threats in domain-specific LLM use, the proposed approach makes theoretical and practical contributions to the field. These two workflows discussed—Zephyr and Deepseek methods—illuminate those selections in text segmenting, embedding models, and retrieval can considerably influence the validity and credibility of AI output. While computational cost and uniformity are favored with the Zephyr model due to fixed-length chunking and top-k retrieval, document organization and MMR-based retrieval make the Deepseek method more logical and context-suitable outputs.

Future work will aim to further refine these pipelines with fine-tuning domain-specific data and investigation into multimodal integration to further increase the system's robustness in real-world humanitarian contexts. Refined measures of hallucination control and ongoing evaluation with next-generation LLM judges will also be essential in making sure AI systems can dependably assist in sensitive psychosocial analysis. Finally, the findings from this research open up the possibilities of more humane and effective AI intervention in humanitarian contexts.